\documentclass[12pt]{article}

\usepackage{amsmath,amsfonts,amssymb,amsthm}
\usepackage{graphicx}

\newtheorem{thm}{Theorem}[section]

\theoremstyle{definition}
\newtheorem{defn}{Definition}[section]

\theoremstyle{remark}

\newcommand{\bx}{\mathbf{x}}
\newcommand{\cX}{\mathcal{X}}

\newcommand{\bD}{\mathbf{D}}
\newcommand{\cD}{\mathcal{D}}

\newcommand{\st}{\mbox{s.t.}}

\newcommand{\fixme}[1]{}

\newcommand{\forblind}[1]{}

\begin{document}

\title{Explicit probabilistic models for databases and networks}

\author{Tijl De Bie\\
University of Bristol, Department of Engineering Mathematics,\\ Bristol, UK\\
tijl.debie@gmail.com}

\maketitle \thispagestyle{empty}

\begin{abstract}
Recent work in data mining and related areas has highlighted the
importance of the statistical assessment of data mining results.
Crucial to this endeavour is the choice of a non-trivial null model
for the data, to which the found patterns can be contrasted. The
most influential null models proposed so far are defined in terms of
invariants of the null distribution. Such null models can be used by
computation intensive randomization approaches in estimating the
statistical significance of data mining results.

Here, we introduce a methodology to construct non-trivial
probabilistic models based on the maximum entropy (MaxEnt)
principle. We show how MaxEnt models allow for the natural
incorporation of prior information. Furthermore, they satisfy a
number of desirable properties of previously introduced
randomization approaches. Lastly, they also have the benefit that
they can be represented explicitly. We argue that our approach can
be used for a variety of data types. However, for concreteness, we
have chosen to demonstrate it in particular for databases and
networks.
\end{abstract}

%-------------------------------------------------------------------------
\section{Introduction}
%-------------------------------------------------------------------------

Data mining practitioners commonly have partial a partial
understanding of the structure of the data investigated. The goal of
the data mining process is then to discover any additional structure
or patterns the data may exhibit. Unfortunately, structure that is
trivially implied by the prior information available is often
overwhelming, and it is hard to design data mining algorithms that
look beyond it. We believe that adequately solving this problem is a
major challenge in current data mining research.

For example, it should not be seen as a surprise that items known to
be frequent in a binary database are jointly part of many
transactions, as this is what should be expected even under a model
of independence. Similarly, in random network theory, a densely
connected community of high degree nodes should probably be
considered less interesting than a similarly tight community among
low degree nodes.% Also, it is well known that the diameter of a
%random Erd\"os-Renyi network is logarithmic in the number of nodes.
%Given that this is a mathematical truth, it should not be
%rediscovered by data mining.

Rather than discovering patterns that are implied by prior
information, data mining is concerned with the discovery from data
of departures from this prior information. To do this, the ability
to formalize prior information is as crucial as the ability to
contrast patterns with this information thus formalized. In this
paper, we focus on the first of these challenges: the task of
designing appropriate models incorporating prior information in data
mining contexts.

We advocate the formalization of prior information using
probabilistic models. This enables one to assess patterns using a
variety of principles rooted in statistics, information theory, and
learning theory. It allows one to formalize the informativeness of
patterns using statistical hypothesis testing---in which case the
probabilistic model is referred to as the null model---, the minimum
description length principle, and generalization arguments
\cite{MSI:02,swap07,webb07,OVK:08,Item06Siebes,Man:08,HOV:09}.

%The statistical assessment of data mining results is increasingly
%recognized as a crucial aspect of the data mining process
%\cite{swap07,webb07,OVK:08,Man:08,HOV:09}. Crucial to this task is
%the definition of a null model. To be useful, this null model must
%non-trivial: it must accurately reflect our prior beliefs about the
%data. Additionally, it must be stated in a form amenable to
%statistical testing.

The most flexible and influential probabilistic models currently
used in data mining research have been defined implicitly in terms
of randomization invariants. Such invariants have been exploited
with success by computationally intensive approaches in estimating
the significance (quantified by the p-value) of data mining results
with respect to the null model they define
\cite{MSI:02,swap07,Man:08,OVK:08,HOV:09}.

Unfortunately, null models defined in terms of invariants cannot be
used to define practical measures of informativeness that can
directly guide the search of data mining algorithms towards the more
interesting ones. To be able to do this, \emph{explicit}
probabilistic models that take prior information into account, are
needed. Applications of models that are defined implicitly in terms
of invariants seem to be limited to \emph{post-hoc} analyses of data
mining results only.

Despite their potential, the development and use of explicit models
is rare in data mining literature, in particular in the study of
databases and networks---the main focus of this paper. Furthermore,
most of the models that have been proposed elsewhere suffer from
serious shortcomings or have a limited applicability (see Discussion
in Sec.~\ref{discussion}).

%One example is the random network model by \cite{ChL:04}, which
%allows one to specify the degree sequence as prior information. A
%related model was considered for binary databases in \cite{swap07}.
%Their potential was recognized in these works, however, as we will
%point out in the Discussion Section, they are also known to suffer
%from serious limitations.

In this paper, we present a methodology for efficiently computing
explicitly representable probabilistic models for general types of
data, able to incorporate non-trivial types of prior information.
Our approach is based on the maximum entropy (MaxEnt) principle
\cite{Jaynes82}. Although the methodology is general, for
concreteness we focus on rectangular databases (binary, integer, or
real-valued) as well as networks (weighted and unweighted, directed
and undirected). We further demonstrate remarkably strong
connections between these MaxEnt models and the aforementioned
randomization approaches for databases and networks.

\paragraph{Outline}
In the remainder of the Introduction, we will first discuss the
maximum entropy principle in general (Sec.~\ref{maxent}). Then we
will discuss how rectangular databases and networks are trivially
represented as a matrix (Sec.~\ref{matrixrep}), and a way to
formalize a common type of prior information for databases and
networks as constraints on that matrix (Sec.~\ref{priorinf}). These
results allow us to study the design of MaxEnt models for matrices
in general, quite independently of the type of data structure it
represents (Sec.~\ref{mainsection}). In
Sec.~\ref{invariancesection}, we relate the MaxEnt distributions to
distributions defined implicitly using swap randomizations. We then
provide experiments demonstrating the scalability of the MaxEnt
modelling approach in a number of settings (Sec.~\ref{experiments}).
In the Discussion (Sec.~\ref{discussion}) we point out relations
with literature and implications the results in this paper may have
for data mining research and applications.

\subsection{The maximum entropy principle}\label{maxent}

Consider the problem of finding a probability distribution $P$ over
the data $\bx\in\cX$ that satisfies a set of linear constraints of
the form:
\begin{eqnarray}
\sum_\bx P(\bx)f_i(\bx)&=&d_i.\label{constraints_f}
\end{eqnarray}
Below, we will show that prior information in data mining can often
be formalized as such. For now, let us focus on the implications of
such a set of constraints on the shape of the probability
distribution. First, note that any probability distribution
satisfies the extra conditions
\begin{eqnarray*}
\sum_\bx P(\bx)=1&,&P(\bx)\geq 0.
\end{eqnarray*}
In general, these constraints will not be sufficient to uniquely
determine the distribution of the data. The most common strategy to
overcome this problem is to search for the distribution that has the
largest entropy subject to these constraints, to which we will refer
as the MaxEnt distribution. Mathematically, it is found as the
solution of:
\begin{eqnarray}
&\max_{P(\bx)}& -\sum_\bx P(\bx)\log P(\bx),\nonumber\\
&\st& \sum_\bx P(\bx)f_i(\bx)=d_i,\ (\forall i)\label{ci_general}\\
&& \sum_\bx P(\bx)=1.\label{c1_general}
\end{eqnarray}

Originally advocated by Jaynes \cite{Jaynes57,Jaynes82} as a
generalization of Laplace's principle of indifference, the MaxEnt
distribution can be defended in a variety of ways. The most common
argument is that any distribution other than the MaxEnt distribution
effectively makes additional assumptions about the data that reduce
the entropy. As making additional assumptions biases the
distribution in undue ways, the MaxEnt distribution is the safest
bet.

A lesser known argument, but not less convincing, is a
game-theoretic one \cite{Top:79}. Assuming that the true data
distribution satisfies the given constraints, it says that the
compression code (e.g. Huffman) designed based on the MaxEnt
distribution minimizes the worst-case expected coding length of a
message coming from the true distribution. Hence, using the MaxEnt
distribution for coding purposes is optimal in a robust minimax
sense.

Besides these motivations for the MaxEnt principle, it is also
relatively easy to compute a MaxEnt model. Indeed, the MaxEnt
optimization problem is convex, and can be solved conveniently using
the method of the Lagrange multipliers. Let us use Lagrange
multiplier $\mu$ for constraint~(\ref{c1_general}) and $\lambda_i$
for constraints~(\ref{ci_general}). The Lagrangian is then equal to:
\begin{eqnarray*}
L(\mu,\lambda_i,P(\bx))&=&-\sum_\bx P(\bx)\log P(\bx)\\
&&+\sum_i\lambda_i\left(\sum_\bx P(\bx)f_i(\bx)-d_i\right)\\
&&+\mu\left(\sum_\bx P(\bx)-1\right).
\end{eqnarray*}
Equating the derivative with respect to $P(\bx)$ to $0$ yields the
optimality conditions:
\begin{eqnarray*}
\log P(\bx) &=& \mu-1+\sum_i \lambda_i f_i(\bx).
\end{eqnarray*}
Hence, the MaxEnt solution belongs to the exponential family of
distributions of the form:
\begin{eqnarray}
P(\bx)&=&\exp{\left(\mu-1+\sum_i\lambda_i
f_i(\bx)\right)}.\label{exponential_family}
\end{eqnarray}

The Lagrange multipliers should be chosen such that the
constraints~(\ref{c1_general}) and (\ref{ci_general}) are satisfied.
These values can be found by minimizing the Lagrangian after
substituting Eq.~(\ref{exponential_family}) for $P(\bx)$, resulting
in the so-called the dual objective function. After some algebra, we
find:
\begin{eqnarray*}
L(\mu,\lambda_i,P(\bx))&=&\sum_\bx\exp\left(\mu-1+\sum_i\lambda_i
f_i(\bx)\right)\\
&& -\mu-\sum_i\lambda_i d_i.
\end{eqnarray*}
This is a smooth and convex function, which can be minimized
efficiently using standard techniques (see Sec.~\ref{sec_lagrange}).

\subsection{Matrix representations of databases and
networks}\label{matrixrep}

In this paper, we will apply the MaxEnt principle for the purpose of
inferring probabilistic models for rectangular databases as well as
networks. Both these data structures can be represented as a matrix,
which we will denote as $\bD$. This is why they are conveniently
discussed within the same paper. Our models will be models for the
random matrix $\bD$, which is directly useful in modelling databases
as well as networks $\bD$ may represent.

For the case of databases, the matrix $\bD$ has $m$ rows and $n$
columns, and $\bD(i,j)$ represents the element at row $i$ and column
$j$. For networks, the matrix $\bD$ represents the $n\times n$
adjacency matrix. It is symmetric for undirected networks and
potentially asymmetric for directed networks. For undirected
networks $\bD(i,j)=\bD(j,i)$ contains the weight of the edge
connecting nodes $i$ and $j$, whereas for directed networks
$\bD(i,j)$ contains the weight of the directed edge from node $i$ to
node $j$. In many cases, self-loops would not be allowed, such that
$\bD(i,i)=0$ for all $i$.

To maintain generality, we will assume that all matrix values belong
to some specified set $\cD\subseteq\Re^+$, i.e.: $\bD(i,j)\in\cD$.
Later we will choose the set $\cD$ to be the set $\{0,1\}$ (to model
binary databases and unweighted networks), the set of positive
integers, or the set of positive reals (to model integer-valued and
real-valued databases and weighted networks). Other choices can be
made, and it is fairly straightforward to adapt the derivations
accordingly.

For notational simplicity, in the subsequent derivations we will
assume that $\cD$ is discrete and countable. However, if $\cD$ is
continuous (such as the set of positive reals) the derivations can
be adapted easily by replacing summations over $\cD$ with
integrals.\fixme{Can we make this more rigorous and sophisticated by
using measure theoretic notation?}

%-------------------------------------------------------------------------
\subsection{Prior information for databases and
networks}\label{priorinf}

For binary databases, it has been argued that row and column sums
(also called row and column marginals) can often be assumed as prior
information. Any pattern that can be explained by referring to row
or column sums in a binary database is then deemed uninteresting.
Previous work has introduced ways to assess the significance of data
mining results based on this assumption \cite{swap07,Man:08,HOV:09}.
These methods were based on the fact that the set of databases with
fixed row and column sums is closed under so-called swaps.

\begin{figure}
\includegraphics[width=\columnwidth]{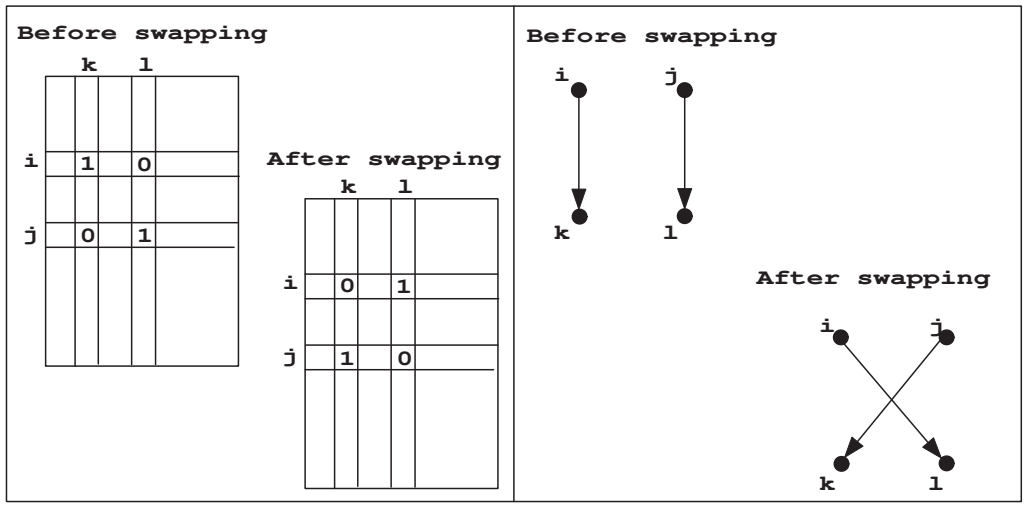}
\caption{The effect of a swap operation to a binary database (left)
and to an unweighted directed network.}\label{swaps}
\end{figure}

Swaps are operations that transform any $2\times 2$ submatrix of the
form $\left(\begin{array}{cc}1&0\\0&1\end{array}\right)$ into
$\left(\begin{array}{cc}0&1\\1&0\end{array}\right)$. Clearly, such
operations leave the row and column sums invariant. Furthermore, it
can be shown that by iteratively applying swap randomizations to
$\bD$, any matrix with the same row and column sums can be obtained.
Thus, randomly applying swaps provides a suitable mechanism to
sample from the set of databases with fixed column and row sums. See
Fig.~\ref{swaps} on the left for a graphical illustration.

The swap operation has later been generalized to deal with
real-valued databases as well \cite{OVK:08}.

Similar ideas have been developed quite independently in the context
of network analysis, and in particular in the search for recurring
network motifs \cite{MSI:02}. There, swaps were applied to the
adjacency matrix, corresponding to a rewiring of the kind depicted
on the right in Fig.~\ref{swaps}. Randomized networks obtained using
a chain of such edge swaps were used to statistically assess the
significance of particular recurring network motifs in biological
networks \cite{MSI:02}.

Note that the sum of row $i$ of an adjacency matrix $\bD$
corresponds to the out-degree of node $i$, whereas the sum of column
$j$ corresponds to the in-degree of node $j$. Clearly, swaps of this
kind leave the in-degree and out-degree of all nodes invariant.
I.e., by using swap operations on networks, one can sample from the
set of all networks with given in-degree and out-degree sequences.

In summary, whether $\bD$ represents a database or a network, the
invariants amount to constraints its row and column sums. The models
we will develop in this paper are based on exactly these invariants,
be it in a somewhat relaxed form: we will assume that the expected
values of the row and column sums are equal to specified values.
Mathematically, this can be expressed as:
\begin{eqnarray*}
\sum_{\bD\in\cD^{m\times n}} P(\bD)\left(\sum_j \bD(i,j)\right)&=&d^r_i,\\
\sum_{\bD\in\cD^{m\times n}} P(\bD)\left(\sum_i
\bD(i,j)\right)&=&d^c_j,
\end{eqnarray*}
where $d_i^r$ is the $i$'th expected row sum and $d_j^c$ the $j$'th
expected column sum. Although they have been developed for binary
databases \cite{swap07} and unweighted networks \cite{MSI:02}, and
later extended to real-valued databases \cite{OVK:08}, we will
explore the consequences of these constraints in broader generality,
for various types of databases and weighted networks.

Importantly, it is easy to verify that these constraints are exactly
of the type of Eq.~(\ref{constraints_f}), such that the MaxEnt
formalism is directly applicable.

%-------------------------------------------------------------------------
\section{MaxEnt Distributions with Given Expected Row and Column
Sums}\label{mainsection}
%-------------------------------------------------------------------------

We have now pointed out that databases as well as networks can be
represented using a matrix over a set $\cD$ of possible values.
Furthermore, commonly used constraints on models for databases as
well as for networks amount to constraining the column and row sums
of this matrix to be constant. Thus, we can first discuss MaxEnt
models for $m\times n$ matrices $\bD$ in general, subject to row and
column sum constraints. Then we will point out particularities and
adjustments to be made for these models to be applicable for
matrices representing databases or networks.

The MaxEnt distribution subject to constraints on the expected row
and column sums is found by solving:
\begin{eqnarray}
&\max_{P(\bD)}& -\sum_{\bD} P(\bD)\log(P(\bD)),\nonumber\\
&\st& \sum_{\bD}
P(\bD)\left(\sum_j \bD(i,j)\right)=d^r_i,\label{ci}\\
&& \sum_{\bD} P(\bD)\left(\sum_i
\bD(i,j)\right)=d^c_j,\label{cj}\\
&& \sum_{\bD} P(\bD)=1.\label{c1}
\end{eqnarray}

The resulting distribution will belong to the exponential family,
and will be of the form of Eq.~(\ref{exponential_family}):
\begin{eqnarray}
P(\bD)&=&\exp\Bigg[\mu-1+\sum_i \lambda_i^r \left(\sum_j \bD(i,j)\right)\nonumber\\
&&+\sum_j \lambda_j^c \left(\sum_i \bD(i,j)\right)\Bigg],\nonumber\\
&=&\exp\Bigg[\mu-1+\sum_{i,j}\bD(i,j)(\lambda_i^r+\lambda_j^c)\Bigg],\nonumber\\
&=&\exp(\mu-1)\prod_{i,j}\exp\left(\bD(i,j)(\lambda^r_i+\lambda^c_j)\right),\label{PD1}
\end{eqnarray}
where $\mu$ is the Lagrange multiplier for constraint (\ref{c1}),
$\lambda^r_i$ are the Lagrange multipliers for constraints
(\ref{ci}), and $\lambda^c_j$ for constraints (\ref{cj}).

The first factor in this expression is a normalization constant, the
value of which can be determined using constraint (\ref{c1}):
\begin{eqnarray*}
\exp(1-\mu) &=&
\sum_{\bD\in\cD^{m\times n}}\prod_{i,j}\exp\left(\bD(i,j)(\lambda^r_i+\lambda^c_j)\right),\\
&=& \prod_{i,j}\sum_{\bD(i,j)\in\cD}\exp\left(\bD(i,j)(\lambda^r_i+\lambda^c_j)\right),\\
&=& \prod_{i,j}Z(\lambda^r_i,\lambda^c_j),
\end{eqnarray*}
where $Z(\lambda^r_i,\lambda^c_j)=
\sum_{\bD(i,j)\in\cD}\exp\left(\bD(i,j)(\lambda^r_i+\lambda^c_j)\right)$.

Plugging this into Equation (\ref{PD1}) yields:
\begin{eqnarray*}
P(\bD) &=&
\prod_{i,j}\frac{1}{Z(\lambda^r_i,\lambda^c_j)}\exp\left(\bD(i,j)(\lambda^r_i+\lambda^c_j)\right),\\
&=& \prod_{i,j}P(\bD(i,j)),
\end{eqnarray*}
where
\begin{eqnarray}
P(\bD(i,j)) =
\frac{1}{Z(\lambda^r_i,\lambda^c_j)}\exp\left(\bD(i,j)(\lambda^r_i+\lambda^c_j)\right)\label{Pindividual}
\end{eqnarray}
is a properly normalized probability distribution for the matrix
element $\bD(i,j)$ at row $i$ and column $j$. Hence, the MaxEnt
model factorizes as a product of independent distributions for the
matrix elements. It is important to stress that we did not impose
independence to start. The independence is a consequence of the
MaxEnt objective.

Various choices for $\cD$ will lead to various distributions, with
appropriate values for the normalization constant
$Z(\lambda^r_i,\lambda^c_j)$. For $\cD$ binary, the MaxEnt
distribution for $\bD$ is a product of independent Bernoulli
distributions with probability of success equal to
$\frac{\exp{(\lambda_i^r+\lambda_j^c)}}{1+\exp{(\lambda_i^r+\lambda_j^c)}}$
for $\bD(i,j)$. For $\cD$ the set of positive integers, the
distribution is a product of independent geometric distributions
with success probability equal to $\exp(\lambda_i^r+\lambda_j^c)$
for $\bD(i,j)$. And for $\cD$ the set of positive reals, the
distribution is a product of independent exponential distributions
with rate parameter equal to $-(\lambda_i^r+\lambda_j^c)$ for
$\bD(i,j)$. This is summarized in Table~\ref{normalization_factors}.
%It can be noted that for $\lambda^t_i+\lambda^c_j$ sufficiently
%small, these first two normalization factors become arbitrarily
%close to each other. For $\lambda^t_i+\lambda^c_j<0$ sufficiently
%close to zero, the last two become arbitrarily close to each other.

\begin{table}
\centering \caption{Three possible domains for the elements of
$\bD$, the corresponding normalization factors in the MaxEnt
distribution $P(\bD)$ for the matrix element $\bD(i,j)$, and the
resulting type of distribution for the matrix
elements.}\label{normalization_factors}
\begin{tabular}{c|c|c}
 $\cD$ & $1/Z(\lambda^r_i,\lambda^c_j)$&Distribution\\\hline
 $\{0,1\}$ & $1/\left(1+\exp(\lambda^r_i+\lambda^c_j)\right)$&Bernoulli\\
 $\mathbb{N}$ & $1-\exp(\lambda^r_i+\lambda^c_j)$&Geometric\\
 $\mathbb{R}^+$ & $-(\lambda^r_i+\lambda^c_j)$&Exponential\\\hline
\end{tabular}
\end{table}

%-------------------------------------------------------------------------
\subsection{MaxEnt models for databases and networks}

The MaxEnt matrix approach is directly applicable for modelling
databases of size $m\times n$, and no further modifications are
required.

Similarly, for directed networks, given the in-degrees and
out-degrees, a MaxEnt model can be found by using the approach
outlined above.

Small modifications are needed for the case of undirected networks
where $\bD(i,j)=\bD(j,i)$. From symmetry, it follows that at the
optimum $\lambda_i^r=\lambda_i^c$ and we can omit the superscripts
$r$ and $c$, such that the solution looks like:
\begin{eqnarray*}
P(\bD)&=&\prod_{i,j}P(\bD(i,j)),\\
P(\bD(i,j))&=&\frac{1}{Z(\lambda_i,\lambda_j)}\exp{\left(\bD(i,j)(\lambda_i+\lambda_j)\right)},
\end{eqnarray*}
where
$Z(\lambda_i,\lambda_j)=\sum_{\bD(i,j)\in\cD}\exp{\left(\bD(i,j)(\lambda_i+\lambda_j)\right)}$.
%Finding the optimal values for $\lambda_i$ is done as outlined in
%Sec~\ref{maxent}, by minimizing the dual optimization problem
%obtained by substituting this solution for $P(\bD)$ into the
%Lagrangian.

Another small modification is needed for networks where self-loops
are not allowed, such that $\bD(i,i)=0$. For $\cD\subseteq\Re^+$
these constraints can be enforced quite easily by requiring
$\sum_\bD P(\bD)\bD(i,i)=0$ for all $i$, which are again constraints
of the form of Eq.~(\ref{constraints_f}). The resulting optimal
distributions are identical in shape to the model with self-loops,
apart from the fact that self-loops receive zero probability and the
Lagrange multipliers would have slightly different values, at the
optimum.

%-------------------------------------------------------------------------
\subsection{Optimizing the Lagrange multipliers}\label{sec_lagrange}

We have now derived the shape of the models $P(\bD)$, expressed in
terms of the Lagrange multipliers (also known as the dual
variables), but we have not yet discussed how to compute the values
of these Lagrange multipliers at the MaxEnt optimum.

In Sec.~\ref{maxent}, we have briefly outlined the general strategy
to do this, as dictated by the theory of convex optimization
\cite{boyd04}: the solution for $P(\bD)$ in terms of the Lagrange
multipliers should be substituted into the Lagrangian of the MaxEnt
optimization problem. The result is a smooth and convex function of
the Lagrange multipliers, and minimizing it with respect to the
Lagrange multipliers yields the optimal values.

Let us investigate what this means for the case of general $m\times
n$ matrices. The number of constraints and hence the number of
Lagrange multipliers is equal to $m+n+1$, which is sublinear in the
size of the data $mn$. The optimal values of the parameters is found
easily using standard methods for unconstrained convex optimization
such as Newton's method or (conjugate) gradient descent, possibly
with a preconditioner \cite{CG,boyd04}. We will report results for
two possible choices in the Experiments Section.

In certain cases, the computational and space complexity can be
further reduced, in particular when the number of distinct values of
$d_i^r$ and of $d_j^c$ are small. Indeed, if $d_i^r=d_k^r$ for
specific $i$ and $k$, the corresponding Lagrange multipliers
$\lambda_i^r$ and $\lambda_k^r$ will be equal as well, reducing the
number of free parameters.

Especially for $\cD=\{0,1\}$, this situation is the rule rather than
the exception. Indeed, when the $d_i^r$ and $d_j^c$ are computed
based on a given database with $m$ rows and $n$ columns, it is
readily observed that both the number of distinct row sums $d_i^r$
and column sums $d_j^c$ are upper bounded by $\mbox{min}(m,n)$,
significantly reducing the complexity when
$\mbox{min}(m,n)\ll\mbox{max}(m,n)$. Furthermore, the number of
different nonzero row sums as well as the number of different
nonzero column sums in a sparse matrix with $s$ nonzero elements is
at most $\sqrt{2s}$.\fixme{Add reference to extremal combinatorics?}
Hence, the number of dual variables is upper bounded by
$2\mbox{min}(m,n,\sqrt{2s})$. Furthermore, this bound can be
sharpened by
$d^r_{\mbox{\scriptsize{max}}}+d^c_{\mbox{\scriptsize{max}}}$ with
$d^r_{\mbox{\scriptsize{max}}}$ and $d^c_{\mbox{\scriptsize{max}}}$
upper bounds on the row and column sums. For $\cD$ the set of
integers, similar bounds can be obtained.

At first sight, it may seem to be a concern that the MaxEnt model is
a product distribution of independent distributions for each
$\bD(i,j)$. However, it should be pointed out that one does not need
to store the value of $\lambda_i^r+\lambda_j^c$ for each pair of $i$
and $j$. Rather, it suffices to store just the $\lambda_i^r$ and
$\lambda_j^c$ to compute the probabilities for any $\bD(i,j)$ in
constant time. Hence, also the space required to store the resulting
model is $O(m+n)$, sublinear in the size of the data.

For each of the models discussed in this paper we will make the code
freely available.

\fixme{Mention the number of variables, perhaps something about
complexity, etc\ldots Refer to Boyd and Vandenberghe?}

\fixme{Mention that I put code online for each of these things.
Also, say something about computational efficiency, and about tricks
to speed things up for binary databases and unweighted networks.}

%-------------------------------------------------------------------------
\section{The Invariance of MaxEnt Matrix Distributions to
$\delta$-Swaps}\label{invariancesection}
%-------------------------------------------------------------------------

We have motivated the use of constraints on the expected row and
column sums by relying on previous work where row and column sums of
a database or a network adjacency matrix was argued to be reasonable
prior information a data mining practitioner may have about the
problem. In this prior work, the authors devised ways to generate
new random data satisfying this prior information, by randomizing
the given database or network using swaps. These swaps allow one to
sample from the uniform distribution over all binary databases with
given row and column sums \cite{swap07}, or from all networks with
given in-degrees and out-degrees \cite{MSI:02}. Later, the swap
operation was generalized to allow randomizing a real-valued data
matrix as well \cite{OVK:08}.

We believe the MaxEnt models introduced in this paper are most
interesting in their own right, being explicitly represented, easy
to compute, and easy to sample random databases or network adjacency
matrices from. Still, it is instructive to point out some relations
between them and the previously proposed swap operations.

%-------------------------------------------------------------------------
\subsection{$\delta$-swaps: a randomization operation on matrices}

First, let us generalize the definition of a swap as follows.
\begin{defn}[$\delta$-swap]
Given an $m\times n$ matrix $\bD$, a $\delta$-swap for rows $i,k$
and columns $j,l$ is the operation that adds a fixed number $\delta$
to $\bD(i,k)$ and $\bD(j,l)$ and subtracts the same number from
$\bD(i,l)$ and $\bD(j,k)$.
\end{defn}

Of course, for a $\delta$-swap to be useful, it must be ensured that
$\bD(i,j)+\delta,\bD(k,j)-\delta,\bD(i,l)-\delta,\bD(k,l)+\delta\in\cD$.
We will refer to such $\delta$-swaps as allowed $\delta$-swaps.
\begin{defn}[Allowed $\delta$-swap]
A $\delta$-swap for rows $i,k$ and columns $j,l$ is said to be
allowed for a given matrix $\bD$ over the domain $\cD$ iff
$\bD(i,j)+\delta,\bD(k,j)-\delta,\bD(i,l)-\delta,\bD(k,l)+\delta\in\cD$.
\end{defn}

Clearly, an allowed $\delta$-swap leaves the row and column sums
invariant. The following Theorem is more interesting.
\begin{thm}
The probability of a matrix $\bD$ under the MaxEnt distribution
subject to equality constraints on the expected row and column sums
is invariant under allowed $\delta$-swaps applied to $\bD$.
\end{thm}
Indeed, it is easily verified from Eq.~(\ref{Pindividual}) that:
\begin{eqnarray*}
&&P(\bD(i,j))\cdot P(\bD(i,l))\\
&&\cdot P(\bD(k,j))\cdot P(\bD(k,l))\\
&=&P(\bD(i,j)+\delta)\cdot P(\bD(i,l)-\delta)\\
&&\cdot P(\bD(k,j)-\delta)\cdot P(\bD(k,l)+\delta)
\end{eqnarray*}
for any $\delta$, rows $i,k$ and columns $j,l$.

This means that for any $2\times 2$ submatrix of $\bD$, adding a
given number to its diagonal and subtracting the same number from
its off-diagonal elements yields the total probability of the data
invariant.

More generally, the MaxEnt distribution assigns the same probability
to any two matrices that have the same row and column sums. This can
be seen from the fact that Eq.~(\ref{PD1}) is independent from $\bD$
as soon as the row and column sums $\sum_j\bD(i,j)$ and
$\sum_i\bD(i,j)$ are given. In statistical terms: the row and column
sums are sufficient statistics of the data $\bD$. We can formalize
this in the following Theorem:
\begin{thm}
The MaxEnt distribution for a matrix $\bD$, conditioned on
constraints on row and column sums of the form
\begin{eqnarray*}
\sum_j\bD(i,j)&=&d^r_i,\\
\sum_i\bD(i,j)&=&d^c_j,
\end{eqnarray*}
denoted as $P(\bD|\sum_j\bD(i,j)=d^r_i,\sum_i\bD(i,j)=d^c_j)$, is
identical to the uniform distribution over all databases satisfying
these constraints.
\end{thm}
This Theorem further strengthens the connection between the uniform
distribution over all matrices with fixed row and column sums, as
sampled from in \cite{swap07,MSI:02,OVK:08} using swap
randomizations, and the MaxEnt distribution.

%-------------------------------------------------------------------------
\subsection{$\delta$-swaps in databases and networks}

The invariants that have been used in computation intensive
approaches for defining null models for databases and networks are
special cases of these more generally applicable $\delta$-swaps.

For binary databases the condition
$\bD(i,j)+\delta,\bD(k,j)-\delta,\bD(i,l)-\delta,\bD(k,l)+\delta\in\cD$
corresponds to the fact that either $\delta=-1$ and
$\bD(i,k;j,l)=\left(\begin{array}{cc}1&0\\0&1\end{array}\right)$, or
$\delta=1$ and
$\bD(i,k;j,l)=\left(\begin{array}{cc}0&1\\1&0\end{array}\right)$.
Then, the $\delta$-swap is identical to a swap in a binary database.
This shows that the MaxEnt distribution of a binary database is
invariant under swaps as defined in \cite{swap07}. For positive
real-valued databases, the $\delta$-swap operations reduce to the
Addition Mask method in \cite{OVK:08}.

Similarly, the edge swap operations on networks can be understood as
$\delta$-swaps with $\delta$ equal to $1$ or $-1$. For networks in
which self-loops are forbidden, swaps with rows $i,k$ and columns
$j,l$ must satisfy $\{i,k\}\cap\{j,l\}=\emptyset$, such that no
self-loops are created. For undirected networks, a $\delta$-swap
operation with rows $i,k$ and columns $j,l$ should always be
accomplished by a symmetric $\delta$-swap with rows $j,l$ and
columns $i,k$, in order to preserve symmetry.

Besides these special cases, allowed $\delta$-swaps, found here as
simple invariants of the MaxEnt distribution, can be used for
randomizing any of the types of databases or networks discussed in
this paper. This being said, it should be reiterated that the
availability of the MaxEnt distribution should make randomizing the
data using $\delta$-swaps unnecessary. Instead one can simply sample
directly from the MaxEnt distribution, thus avoiding the
computational cost and potential convergence problems faced in
randomizing the data. An exception would be if it is crucial that
the row and column sums are preserved exactly rather than in
expectation.

\fixme{Somewhere we should have a theorem that given independence,
the only model that satisfies this as well as constraints on the
expected row and column sums is the MaxEnt model.}

%-------------------------------------------------------------------------
\section{Experiments}\label{experiments}
%-------------------------------------------------------------------------

In this Section we present experiments that illustrate the
computational cost of the MaxEnt modelling approach on a number of
problems. All experiments were done on a 2GHz Pentium Centrino with
1GB Memory.

%-------------------------------------------------------------------------
\subsection{Modelling binary databases}

We report empirical results on four databases: two textual datasets,
turned into databases by considering words as items and documents as
transactions, and two other databases commonly used for evaluation
purposes.
\begin{description}
\item[ICDM] All ICDM abstracts until 2007, where each abstract is
represented by as a transaction and words are items. Stop words have
been removed, and stemming performed.
\item[Mushroom] A publicly available item-transaction dataset \cite{UCI}.
\item[Pubmed] All Pubmed abstracts retrieved by querying with the search
query ``data mining", after stop word removal and stemming.
\item[Retail] A dataset about transactions in a Belgian supermarket
store, where transactions and items have been anonymized
\cite{retail}.
\end{description}
Some statistics are gathered in Table~(\ref{datasets}). The Table
also mentions support thresholds used for some of the experiments
reported below, and the numbers of closed itemsets satisfying these
support thresholds.

\begin{table}\caption{Some statistics for the databases investigated.}\label{datasets}
\begin{center}
\begin{tabular}{|r|cccc|}\hline
& \# items & \# tids & support & \# closeds \\\hline%
ICDM & 4,976 & 859 & 5 & 365,732 \\%
Mushroom & 120 & 8,124 & 812 & 6,298 \\%
Pubmed & 12,661 & 1,683 & 10 & 1,249,913 \\%
Retail & 16,470 & 88,162 & 8 & 191,088 \\\hline%
\end{tabular}
\end{center}
\end{table}

\paragraph{Fitting the MaxEnt model}
The method we used to fit the model is a preconditioned gradient
descent method with Jacobi preconditioner (see e.g. \cite{CG}),
implemented in C++. It is quite conceivable that more sophisticated
methods will lead to significant further speedups, but this one is
particularly easy to implement.

\begin{figure}
\begin{center}
\includegraphics[width=\columnwidth]{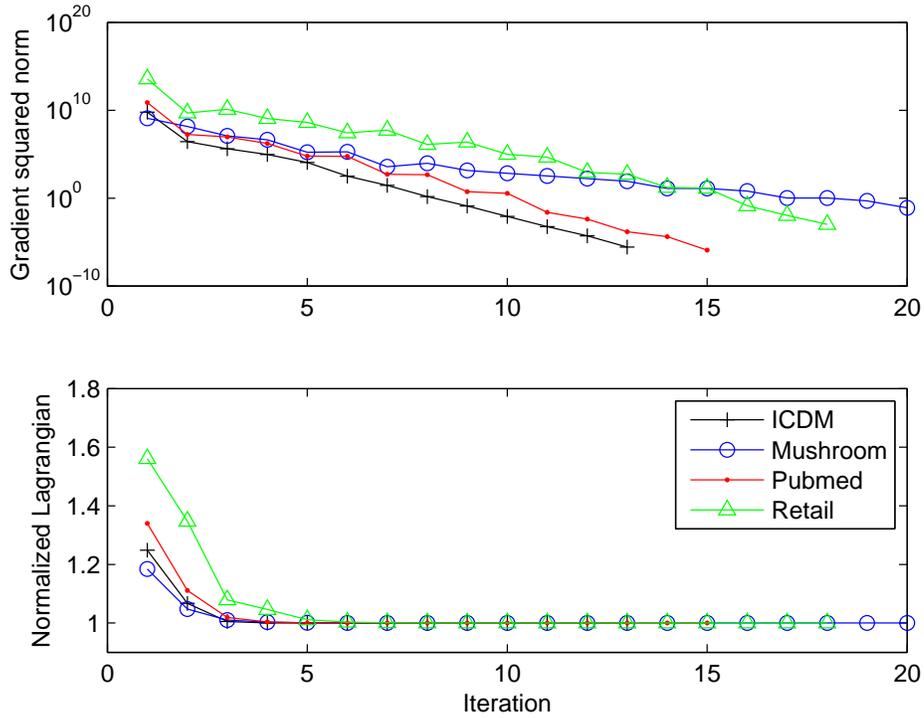}
\end{center}
\caption{Top: the squared norm of the gradient on a logarithmic
scale as a function of the iteration number, plotted for four
databases: ICDM abstracts, Mushroom, Pubmed abstracts, and Retail.
This plot shows the exponential decrease of the gradient of the dual
optimization problem. In the second plot, the convergence of the
Lagrange dual is shown for the same
databases.}\label{convergence_analysis}
\end{figure}

To illustrate the speed to compute the MaxEnt distribution,
Fig.~\ref{convergence_analysis} shows plots of the convergence of
the squared norm of the gradient to zero, for the first 20
iterations. The initial value for all dual variables was chosen to
be equal to $0$. Noting the logarithmic vertical axis, the
convergence appears clearly exponential. The lower plot in
Fig.~\ref{convergence_analysis} shows the convergence of the dual
objective to its minimum over the iterations, clearly a very fast
convergence in just a few iterations.

%It is instructive to note that the gradient vector (see
%Eq.~(\ref{grad1},\ref{grad2})) contains as its elements the
%difference between the empirical marginal probabilities $p_{\ttid}$
%and $p_{\titem}$ on the one hand, and their counterparts under the
%probabilistic model with parameters at the present iteration on the
%other hand. This means that as soon as the gradient is equal to
%zero, the primal constraints Eq.~(\ref{itemfreqs},\ref{tidfreqs})
%are satisfied.
%
%This means that the norm of the gradient, normalized by the number
%of items plus the number of transactions, provides for a suitable
%stopping criterion.
In all our experiments we stopped the iterations as soon as this
normalized squared norm became smaller than $10^{-12}$, which is
close to machine accuracy and accurate enough for all practical
purposes. The number of iterations required and the overall
computation time are summarized in Table~\ref{computation}.

\begin{table}\caption{The number of iterations required, and the
computation time in seconds to fit the probabilistic model to the
data.}\label{computation}
\begin{center}
\begin{tabular}{|r|cc|}\hline
& \# iterations & time (s)\\\hline%
ICDM & 13 & 0.35 \\%
Mushroom & 37 & 0.02 \\%
Pubmed & 15 & 1.24 \\%
Retail & 18 & 2.80 \\\hline%
\end{tabular}
\end{center}
\end{table}

\begin{figure}
\begin{center}
\includegraphics[width=\columnwidth]{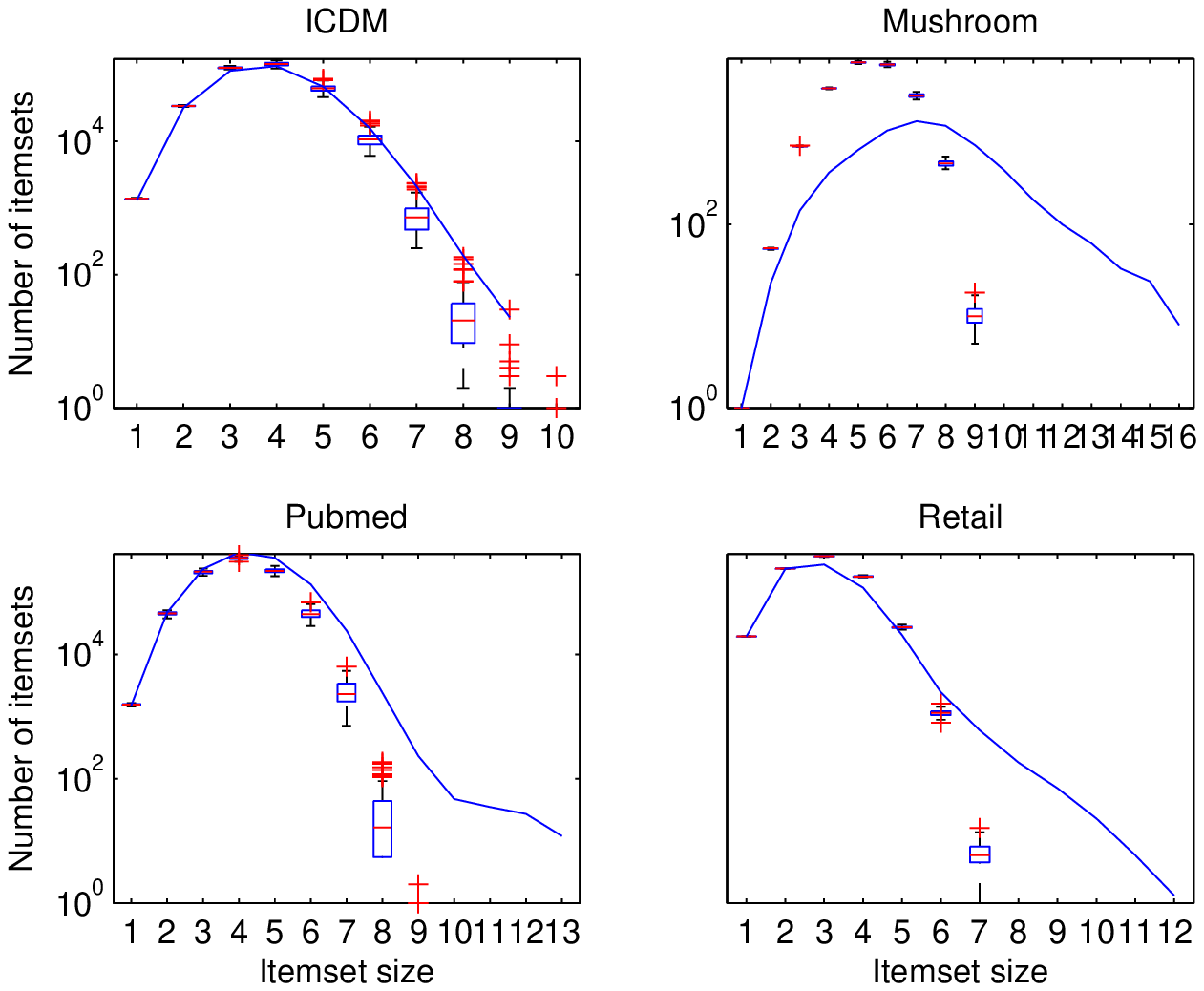}
\end{center}
\caption{For the four datasets under investigation, these plots show
the number of closed itemsets on a logarithmic scale, as a function
of their size. Additionally, box plots are shown for the number of
closed itemsets as a function of size found on 100 randomized
datasets, based the MaxEnt
distribution.}\label{randomization_testing}
\end{figure}

\paragraph{Assessing data mining results}
Here we illustrate the use of the MaxEnt model for assessing data
mining results in the same spirit as \cite{swap07}.
Figure~\ref{randomization_testing} plots the number of closed
itemsets retrieved on the original data. Additionally, it shows box
plots for the results obtained on randomly sampled databases from
the MaxEnt model with expected row sums and column sums constrained
to be equal to their values on the original data. If desired, one
could extract one global measure from these results, as in
\cite{swap07}, and compute an empirical p-value by comparing that
measure obtained on the actual data with the result on the
randomized versions. However, the plots given here do not force one
to make such a choice, and they still give a good idea of the
patterns contained in the datasets.

%As a side-note, we remark that the number of closed itemsets in the
%Retail dataset is indeed smaller than what is typically obtained in
%a randomly generated dataset, as observed in \cite{swap07}. However,
%this seems to be attributable by the fact that large tiles, only
%present in the actual dataset, are shattered in the randomized
%versions, thus leading to a larger number of smaller ones.

Note that the possible applications of the MaxEnt model will likely
reach further than the assessment of data mining results. However,
this is beyond the scope of the current paper, and we will get back
to this in the Discussion Section.

%-------------------------------------------------------------------------
\subsection{Modelling networks}

To evaluate the MaxEnt model for networks, we artificially generated
power-law (weighted) degree distributions for networks of various
sizes between $n=10$ and $n=10^6$ nodes, with a power-law exponent
of $2.5$. I.e., for each $n$ we sampled $n$ expected (weighted)
degrees $d_i$ from the distribution $P(d_i)\sim d_i^{-2.5}$. A
power-law degree distribution with this exponent is often observed
in realistic networks \cite{New:03}, so we believe this is a
representative set of examples. However, non-reported experiments on
other realistic types of networks yield qualitatively similar
results. For each of these degree distributions, we fitted four
different types of undirected networks: unweighted networks with and
without self-loops, and positive integer-valued weighted networks
with and without self-loops.

\begin{figure}
\begin{center}
\includegraphics[width=0.5\columnwidth]{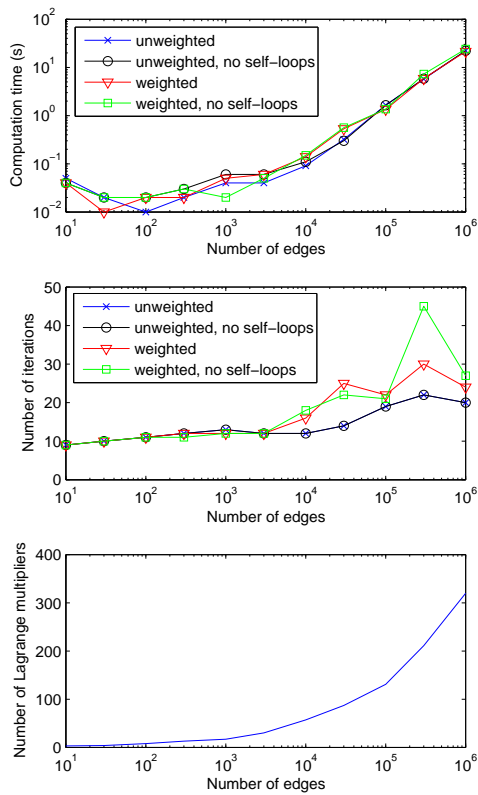}
\end{center}
\caption{Top: The computation time as a function of the size number
of nodes in the network (left). A marker $\times$ is used for the
unweighted model with self-loops, $\circ$ for the unweighted model
without self-loops, $\triangledown$ for the weighted model with
self-loops, and $\Box$ for the weighted model without self-loops.
Note the log-log scale. Middle: The number of iterations required by
the Newton algorithm before convergence. Note the log-scale on the
horizontal axis. Bottom: the number of Lagrange multipliers (i.e.
the number of variables in the dual of the MaxEnt optimization
problem) for the degree sequences investigated, as a function of the
network size. Again, note the log-scale on the horizontal
axis.}\label{time_newton}
\end{figure}

To fit the MaxEnt models for networks we made use of Newton's
method, which we implemented in MATLAB. As can be seen from
Fig.~\ref{time_newton}, the computation time was under 30 seconds
even for the largest network with $10^6$ nodes. The number of Newton
iterations is lower than $50$ for all models and degree
distributions considered. \fixme{Note that the increase in
computation time seems linear, as it increases by 3 order of
magnitude for the last 3 orders of magnitude increase in the network
size. Can something like this even be proven for power-law
distributions??}

This fast performance can be achieved thanks to the fact that the
number of different degrees observed in the degree distribution is
typically much smaller than the size of the network (see Discussion
in Sec.~\ref{sec_lagrange}). The bottom graph in
Fig.~\ref{time_newton}, showing the number of Lagrange multipliers
as a function of the network size supports this. The memory
requirements remain well under control for the same reasons.

It should be pointed out that in the worst case for dense or for
weighted networks (and in particular for real-valued weights), the
number of distinct expected weighted degrees and hence the number of
Lagrange multipliers can be as large as the number of nodes $n$.
This would make it much harder to use off-the-shelf optimization
tools for $n$ much larger than $1000$. However, the problem can be
made tractable again if it is acceptable to approximate the expected
weighted degrees by grouping subsets of them together into bins, and
replacing their values by a bin average. In this way the number of
Lagrange multipliers can be brought below $1000$. We postpone a full
discussion of this and other strategies to a later paper.

\fixme{Only include experiments of computation time of fitting the
model, and maybe also include some temperature plots. Also: a plot
of a network or perhaps some statistics about the network such as
diameter etc, and a randomized version of the network as well
(plotted using a nice graph visualization package), and the same
statistics for a number of randomized networks.}

%-------------------------------------------------------------------------
\section{Discussion}\label{discussion}
%-------------------------------------------------------------------------

We will first discuss how some existing explicit models are related
to particular cases of the MaxEnt models introduced in this paper.
Then, we will discuss the implications for data mining of the
availability explicit models.

%-------------------------------------------------------------------------
\subsection{Connections with literature}\label{literature}

Interestingly, the MaxEnt model for binary matrices introduced in
this paper is formally identical to the Rasch model, known from
psychometrics \cite{Rasch61}. This model was introduced to model the
performance of individuals (rows) to questions (columns). The matrix
elements indicate which questions were answered correctly or
incorrectly for each individual. The Lagrange dual variables are
interpreted as persons' abilities for the row variables
$\lambda_i^r$, and questions' difficulties $\lambda_j^c$.
Remarkably, the model was not derived from the MaxEnt principle but
stated directly.

A similar connection exists with the so-called $p^*$ models from
social network analysis \cite{RPK:07}. Although motivated
differently, the $p_1$ model in particular is formally identical to
our MaxEnt model for unweighted networks.

Thus, the present paper provides an additional way to look at these
widely used models from psychometrics and social network analysis.
Furthermore, as we have shown, the MaxEnt approach suggests
generalizations, in particular towards non-binary databases and
weighted networks.

Another connection is to prior work on random network models for
networks with prescribed degree sequences (see \cite{New:03} and
references therein). The most similar model to the ones discussed in
this paper is the one from \cite{ChL:04}. In this paper, the authors
propose to assume that edge occurrences are independent, with each
edge probability proportional to the product of the degrees of the
pair of nodes considered. In the notation of the present paper:
\begin{eqnarray*}
P(\bD)=\prod_{i,j}P(\bD(i,j))&\mbox{with}&
P(\bD(i,j))=\frac{d_id_j}{s},
\end{eqnarray*}
where $s=\sum_i d_i$. Also for this model the constraints on the
expected row and column sums are satisfied.

It would be too easy to simply dismiss this model by stating that
among all distributions satisfying the expected row and column sum
constraints, it is not the maximal entropy one, such that it is
biased in some sense. However, this drawback can be made more
tangible: the model represents a probability distribution only if
$\max_{i,j}d_id_j\leq s$, which is by no means true in all practical
applications, in particular in power-law graphs. This shortcoming is
a symptom of a bias of this model: it disproportionally favours
connections between pairs of nodes both of high degree, such that
for nodes of too high degrees the edge `probability' suggested
becomes larger than $1$. A brief remark considering a similar model
for binary databases was made in \cite{swap07}, where it was
dismissed by the authors on
similar grounds. %However, it is not clear if these authors were
%aware of the relation to \cite{ChL:04}.

%-------------------------------------------------------------------------
\subsection{Relevance to data mining}

The implications of the ability to construct explicit models for
databases and networks are vast. It suffices to look at an area of
data mining that has been using data models for a long time in the
search for patterns: string analysis.

In the search for patterns in strings, it is common to adopt Markov
models of an appropriate order as a background model. \fixme{Note
that Markov models with transition probability distributions in the
exponential family are MaxEnt models subject to constraints on the
expected transition frequencies!} Then, patterns that somehow depart
from the expected under this background model are reported as
potentially interesting. For example, a comprehensive survey about
motif discovery algorithms \cite{Tompa05} lists various algorithms
that model DNA sequences using Markov models, after which motifs are
scored based on various information theoretic or statistical
measures, depending on the method. One notable example is
MotifSampler \cite{motifsampler}.
%are The Improbizer \cite{improbizer}, and Oligodyad Analysis \cite{oligodyad}
Certainly, randomization strategies have been and continue to be
used successfully for testing the significance of string patterns.
However, it should be emphasized that the methods for discovering
string patterns that contrast with a background model have become
possible only thanks to the availability of that background model in
an explicit form.

Therefore, we believe it is to be expected that the availability of
explicit models for databases and networks will enable the
introduction of new measures of interestingness for patterns found
in such data, be they recurring itemsets or network motifs, other
local patterns, or more global patterns such as correlations between
items or transactions, the clustering coefficient of a network, and
so on.

We also wish to stress that the MaxEnt modelling strategy is
applicable for other types of constraints as well. For example, it
would be possible to add constraints for the support of specific
itemsets, or the presence of certain cliques in a network, etc. Such
modifications would be in line with the alternative randomization
strategies suggested in \cite{HOV:09}. It is also possible to allow
matrix elements to become negative, if further constraints for
example on the variance are introduced (the resulting MaxEnt
distribution would then be a product of normal distributions).
Furthermore, the strategy can be applied to other data types as
well.

%Thus, besides advocating MaxEnt models as having significant
%advantages over randomization approaches and existing explicit
%models (see Sec.~\ref{literature}), we hope this work will help in
%promoting the use of explicit models in data mining.

%-------------------------------------------------------------------------
\section{Conclusions}
%-------------------------------------------------------------------------

In recent years, a significant amount of data mining research has
been devoted to the statistical assessment of data mining results.
The strategies used for this purpose have often been based on
randomization testing and related ideas. The use of randomization
strategies was motivated by the lack of explicit models for the
data.

We have shown that in a wide variety of cases it is possible to
construct explicitly represented distributions for the data. The
modelling approach we have suggested is based on the maximum entropy
principle. We have illustrated that fitting maximum entropy
distributions often boils down to well-posed convex optimization
problems. In the experiments, we have demonstrated how the MaxEnt
model can be fitted extremely efficiently to large databases and
networks in a matter of seconds on a basic laptop computer.

This newfound ability holds several promises for the domain of data
mining. Most trivially, the assessment of data mining results may be
facilitated in certain cases, as sampling random data instances from
an explicit model may be easier than using MCMC sampling techniques
from a model defined in terms of invariants of the distribution.
More importantly, we believe that our results may spark the creation
of new measures of informativeness for patterns discovered from
data, in particular for databases and for networks.

\forblind{\paragraph{Acknowledgements.} This work is supported by
the EPSRC grant EP/G056447/1. The author is grateful to Bart
Goethals for interesting discussions relating to this work.}

\fixme{Also mention the possible application of this approach to
other data types, such as strings / time series / \ldots?}

\bibliographystyle{latex8}
\bibliography{biblio}

%\appendix
%%-------------------------------------------------------------------------
%\section{Proofs}
%%-------------------------------------------------------------------------

\end{document}